\documentclass[letterpaper]{article} 
\usepackage{aaai24}
\usepackage{times}  
\usepackage{helvet}  
\usepackage{courier}  
\usepackage[hyphens]{url}  
\usepackage{graphicx} 
\usepackage{natbib}  
\usepackage{caption} 
\usepackage{algorithm}
\usepackage{algorithmic}
\usepackage{adjustbox}
\usepackage{amsmath}
\usepackage{amssymb}
\usepackage{newfloat}
\usepackage{listings}

\usepackage{tabularray}
\usepackage{multirow}
\usepackage{rotating}
\usepackage{svg}

\DeclareCaptionStyle{ruled}{labelfont=normalfont,labelsep=colon,strut=off} 
\floatstyle{ruled}
\newfloat{listing}{tb}{lst}{}
\floatname{listing}{Listing}
\nocopyright 

\title{Distraction-free Embeddings for Robust VQA}
\author {
    Atharvan Dogra\textsuperscript{\rm 1, \rm 2},
    Deeksha Varshney\textsuperscript{\rm 3},
    Ashwin Kalyan\textsuperscript{\rm 4},
    Ameet Deshpande\textsuperscript{\rm 5},
    Neeraj Kumar\textsuperscript{\rm 1}
}
\affiliations {
    \textsuperscript{\rm 1}Thapar University,
    \textsuperscript{\rm 2}IIT Madras,
    \textsuperscript{\rm 3}IIT Patna,\\
    \textsuperscript{\rm 4}The Allen Institute for AI,
    \textsuperscript{\rm 5}Princeton University\\
    atharvandogra007@gmail.com, deeksha.varshney2695@gmail.com
}

\usepackage{bibentry}

\begin{document}
\maketitle

\begin{abstract}
The generation of effective latent representations and their subsequent refinement to incorporate precise information is an essential prerequisite for Vision-Language Understanding (VLU) tasks such as Video Question Answering (VQA). However, most existing methods for VLU focus on sparsely sampling or fine-graining the input information (e.g., sampling a sparse set of frames or text tokens), or adding external knowledge.
We present a novel \emph{``DRAX: Distraction Removal and Attended Cross-Alignment"} method to rid our cross-modal representations of distractors in the latent space. We do not exclusively confine the perception of any input information from various modalities but instead use an attention-guided distraction removal method to increase focus on task-relevant information in latent embeddings. \emph{DRAX} also ensures semantic alignment of embeddings during cross-modal fusions. We evaluate our approach on a challenging benchmark (SUTD-TrafficQA dataset), testing the framework’s abilities for feature and event queries, temporal relation understanding, forecasting, hypothesis, and causal analysis through extensive experiments.
\end{abstract}

\section{Introduction}

\begin{figure*}[th!]
  \centering
  \includegraphics[width=0.9\textwidth]{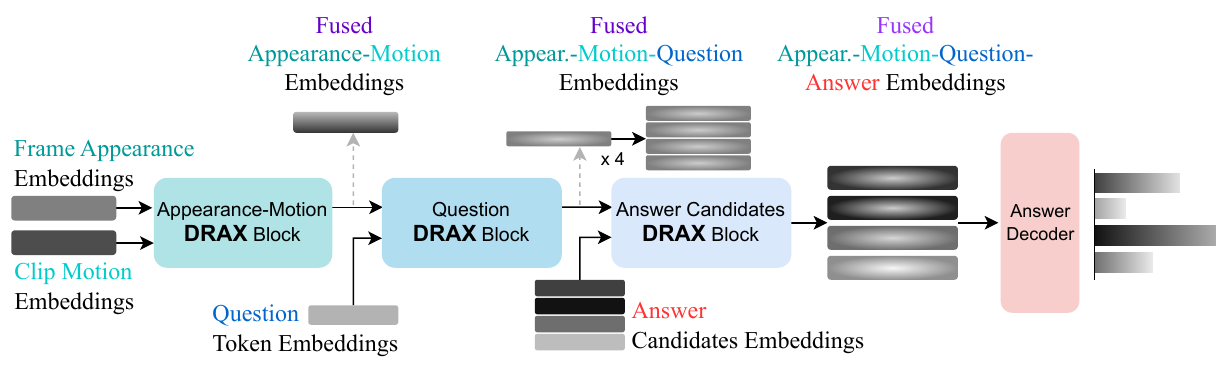}
  \caption{Structured Hierarchy of Input Modalities to the Distraction Removal and Attended Cross Alignment (\textbf{DRAX}) blocks. Each solid (or gradient) rectangle represents a set of vectors, until after the output of Answer Candidates DRAX block where the set of vectors is reduced to a single to fuse with answer candidates.} \label{fig:1}
\end{figure*}

The process of comprehending an environment relies on receiving specific sensory inputs. As the number and diversity of these inputs increase, our level of cognition also tends to rise. Humans, for example, possess primary sensory inputs such as sight, sound, touch, smell, and taste, which contribute to our understanding of the world. However, as the quantity of these inputs grows, it leads to an \emph{information overload} \cite{gross1964managing}, and it becomes increasingly challenging to discern which information is essential and should be prioritized and which can be disregarded. 

In the context of multi-modal systems that combine vision and language, achieving a relational understanding between the two modes of input is crucial for various tasks. Significant advancements have been made in areas like text-to-video retrieval \cite{xu2016msr,krishna2017dense,rohrbach2015dataset,lei2020tvr,li2020hero}, video-text matching \cite{ali2022video}, text-based video moment retrieval \cite{krishna2017dense,lei2020tvr,anne2017localizing, gao2017tall}, video captioning \cite{xu2016msr,rohrbach2015dataset,wang2019vatex, zhou2018towards} and video question answering \cite{jang2017tgif,xu2017video,lei-etal-2018-tvqa,lei2019tvqa+}.

One of the tasks in Video-Language understanding and the task of our focus, Video Question Answering, has seen developments both with \cite{anderson2018bottom,agrawal2018don,ben2017mutan} and without \cite{le2020hierarchical} attention-based approaches. The fundamental concept behind this task involves generating global feature representations for both images and questions, merging them into a shared space \cite{gao2016compact, kim2017hadamard} through fusion, and feeding the fused features into a classifier \cite{anderson2018bottom} to generate answer probabilities. Given that questions can be diverse and multidirectional \cite{Xu_2021_CVPR}, video question-answering systems must encode representations for crucial properties within a video, such as temporal relations \cite{jang2017tgif, li2019beyond, zhou2017temporalrelation}, and relations among different modalities \cite{lu2019vilbert}. These systems also need to learn to focus on relevant and informative features and relationships. Cross-modal attention techniques have proven effective in capturing intricate relationships among different modalities \cite{tan2019lxmert, lei2021less}.

However, previous such methods \cite{le2020hierarchical,lu2019vilbert,tan2019lxmert} operating on dense video and text features often suffered from the inclusion of excessive and irrelevant information. Methods like \emph{ClipBERT} \cite{lei2021less} and \emph{Eclipse} \cite{Xu_2021_CVPR} have tackled this issue by sparse sampling of frames and, to an extreme, selecting one frame at a time over multiple steps. \emph{ClipBERT} also address domain disconnection due to extracting offline features, as seen in \emph{ECLIPSE} \cite{Xu_2021_CVPR} and \emph{HCRN} \cite{le2020hierarchical}, and incorporates a cross-attention mechanism for frames and text tokens, enabling an end-to-end learning process that encompasses even the pixel-level details of video frames.

Several works have successfully addressed the need for high-quality datasets. For example, \cite{lei-etal-2018-tvqa} released \emph{TVQA} leveraging the abundance of visual language data in television shows, covering domains such as crime, medical, and sitcom, and providing dialogue textual captions.  Another notable development is the introduction of \emph{DocVQA} \cite{mathew2021docvqa}, emphasizing that reading systems should utilize cues like layout, non-textual elements, and style. This represents a departure from scene text visual question answering \cite{biten2019scene, singh2019vqa} and generic visual question answering \cite{goyal2017making} approaches.


Newer video-language understanding works have focussed on bringing in more modalities by processing raw inputs \cite{liu2021hit}, extracting discrete tokens \cite{fu2021violet} from the input video, end-to-end training of image encoders \cite{NEURIPS2021_c6d4eb15}, and learning video representations through extensive pre-training \cite{li2023blip2}. 
All discussed methods have had their primary focus on the input modalities or making systems end-to-end or random sampling \cite{lei2021less}, while no work until now, to the best of our knowledge, has looked into making the latent embeddings more robust to focus on task-specific elements. 
We propose a simple method in this new direction of removing irrelevant information from the latent representations and making them ``distraction-free".
To show the efficiency of our proposal in generating effective latent representations, we design a simple framework keeping the offline feature extraction method, and our comparisons are limited to similar works. 

The \emph{SUTD TrafficQA} dataset \cite{Xu_2021_CVPR} (described in ix A) provides a comprehensive set of tasks to evaluate the framework's (1) feature and event recognition capabilities, (2) understanding of temporal relations, (3) judge the capabilities on hypothetical situations, (4) enables forecasting capabilities in the framework and (5) perform causal analysis. 

Our framework is structured to condition the appearance, motion, and linguistic features hierarchically \cite{le2020hierarchical}, using self and cross attention, including cross-attended vector transformations for multi-modal semantic alignment and a guided distraction masking technique, which acts as a filter before producing cross attended vectors. Guided masking helps to focus on relevant information and ignore distractions, which roughly corresponds to the notion of \emph{attention control} in psychology \cite{james1890principles}.

Major contributions of this work are as follows:
\begin{enumerate}
\item We propose a novel approach ``Distraction Removal and Attended Cross Alignment (DRAX)" that identifies and removes distractors from the embeddings and semantically aligns embeddings for fusing multi-modal representations while conditioning modalities in a hierarchical fashion.

\item We incorporate distraction masking in both cross-attention and semantic alignment during fusion which refines the embeddings during all cross-modal interactions.

\item We perform an extensive study on the driving scenes to display the effectiveness of our method in understanding event queries, temporal relation understanding, forecasting, hypothesis and causal analysis through the tasks provided in the dataset SUTD-TrafficQA.
\end{enumerate}
\begin{figure*}
  \centering
  \includegraphics[width=0.95\textwidth]{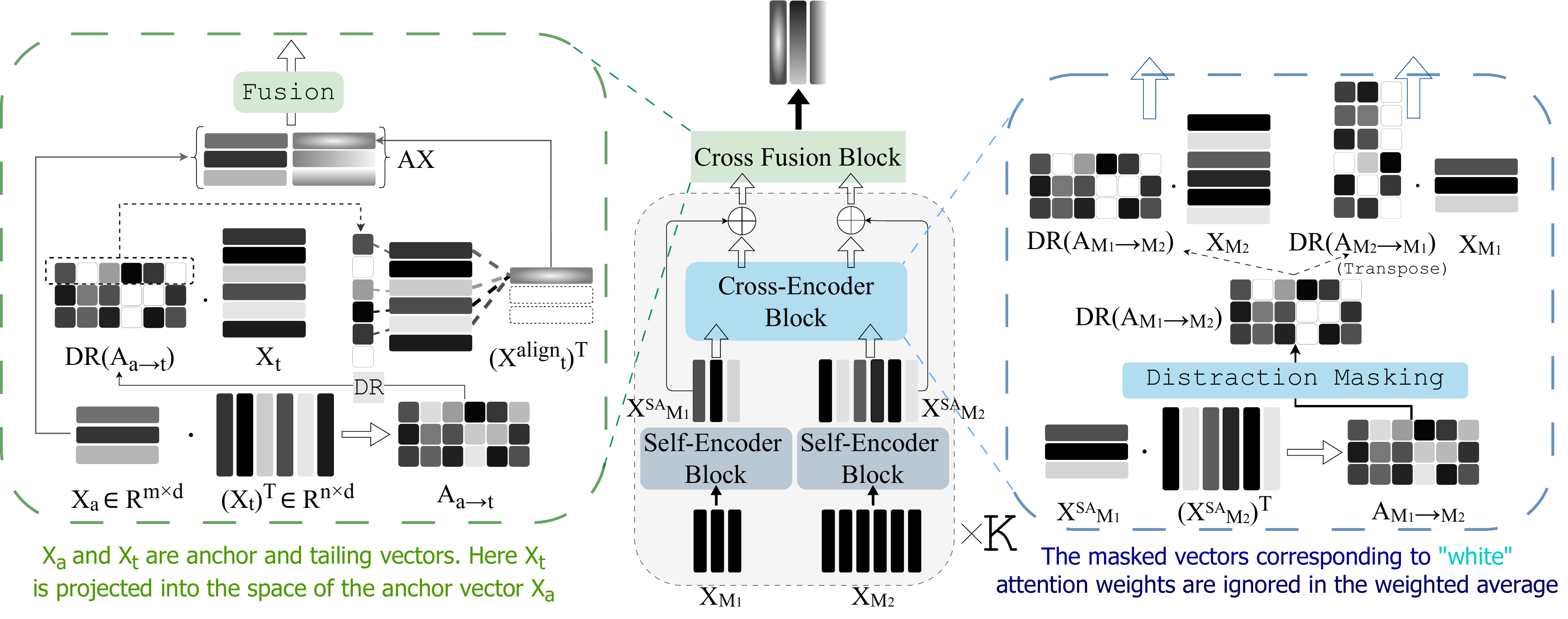}
  \caption{Architecture of the \textbf{DRAX}: Distraction Removal and Attended Cross Alignment Block framework. Distraction Removal ($DR$) and Attended Cross Alignment ($AX$) functions are shown in the zoomed-in views of Cross-Encoder and Cross-Fusion blocks. NOTE: Vectors displayed in vertical form are not $X_M^T$ unless specifically mentioned. DR $\simeq$ Distraction Masking.}\label{fig:2} 
\end{figure*}

\section{DRAX Framework}
The task for the network $\mathcal{F}_{\theta}$ here is to select an answer out of the given answer candidates in set $\mathcal{A}$. In Equation (\ref{eq:1}), $q$ and $\mathcal{V}$ represent the question and visual information, respectively. In this work, we are limiting our application to multi-choice answers with four answer candidates.

\begin{equation}
    \widetilde{a} = {\underset{a \in \mathcal{A}}{\text{argmax}}} \mathcal{F}_{\theta} (a | q, \mathcal{V})
    \label{eq:1}
\end{equation}

The component structure of our VQA system starts with self-attention encoders \cite{NIPS2017_3f5ee243} followed by a cross-modality encoder \cite{tan2019lxmert}, which employs a dynamic attention-guided distraction masking mechanism. Lastly, using cross-attended vector-space transformation, we fuse two vector embeddings, either from individual modalities or previously fused vectors, and a new modality input vector (Figure \ref{fig:1}). All the above components are applied to $\mathcal{V}$ video feature vectors (motion and appearance), and the resulting fused representation vectors are added with linguistic context from question feature vectors. To generate probabilities for answer candidates, individual candidate embeddings are fused with representation from the previous step and are fed to the decoder. $\mathcal{F}_{\theta}$ computes logits, and the answer candidate with the highest probability is selected using the $argmax$ operation.
Our system is structured hierarchically \cite{ijcai2017p492, ijcai2018p513, le2020hierarchical} as shown in Figure \ref{fig:1} where each modality is given as input during different stages of the hierarchy to refine feature vectors.

In our approach, we partition the video input into $\mathcal{C}$ clips of equal lengths and extract $\mathcal{N}$ frames uniformly from the entire video \cite{le2020hierarchical}, which form the  $\mathcal{M}$ and $\mathcal{E}$ vector sets, respectively. Below we describe the different modality inputs which are considered independent in our work:
\smallskip
\subsubsection{Appearance Features} These are represented by the sequence of feature vectors $\{\epsilon_i | \mathcal{E} \in \mathbb{R}^{512}\}_{i=1}^\mathcal{N}$, which correspond to the $\mathcal{N}$ frames. In our implementation, we calculate these features using the CNN-based ResNet-18 model \cite{he2015deep}.

\smallskip

\subsubsection{Motion Features} The motion features $\{m_{i} | \mathcal{M} \in \mathbb{R}^{2048}\}_{i=1}^{\mathcal{C}}$ are a sequence of feature vectors that represent the $\mathcal{C}$ clips. They capture motion information with the intuition of encoding temporal relations and dynamic context among frames or clips. To extract these motion features, we utilize the ResNeXt-101 \cite{xie2017aggregated}.

\smallskip


\subsubsection{Linguistic Representation} The \emph{questions} and \emph{answer candidates} are transformed into $\mathbb{R}^{300}$ space vectors using GloVe word embeddings \cite{pennington-etal-2014-glove}. These linguistic features serve as a single modality. However, the question $\mathcal{Q} \in \mathbb{R}^{d_{q}}$ and answer candidates $\mathcal{A} \in \mathbb{R}^{d_{a}}$ representations are fused separately at different stages in the hierarchical process and operated on individually by the DRAX block. Here $d_q$ and $d_a$ are $300$.
\smallskip

Other components inputted into the encoder, ``added" to the above-described embeddings or the fused embeddings $X_{feat}$ produced in between the hierarchical blocks, are:
\begin{align*}
X_M = [x_{cls} | X_{feat} ] + x_{pos}
\end{align*}
\noindent \textbf{CLS and Positional Encoding} \quad We extend the input sequences at each step of the input hierarchy by appending CLS tokens $x_{cls}$ \cite{devlin2019bert} which capture the overall sequence information and facilitate information transfer between embeddings. Furthermore, we add positional encodings $x_{pos}$ to account for the position-agnostic behavior of the attention mechanism. Understanding the position of elements is crucial for both linguistic and vision comprehension tasks. In contrast to existing literature, we utilize sinusoidal encodings \cite{NIPS2017_3f5ee243} for linguistic embeddings, and we adopt learnable 1D positional encoding for motion and appearance, and the previously fused embedding vectors, inspired by ViT \cite{dosovitskiy2021image} and CrossViT \cite{chen2021crossvit}.

\subsection{Encoders}
In our framework, the encoder stack comprises two separate single-input self-attention encoders along with a cross-modality encoder for the cross-attention operation. 

\subsubsection{Background:} \textbf{Attention} \; The basic idea of attention \cite{bahdanau2016neural, pmlr-v37-xuc15}, involves retrieving information from a set of context vectors ${y_j}$ that are ``\emph{relevant}" to a query vector $x$. This retrieval is accomplished by calculating matching scores $a_i$ between the query vector $x$ and each context vector $y_i$. These matching scores are then normalized using $softmax$:
\begin{equation}
    \begin{aligned}
        a_i = \text{score}(x,y_i);
        \quad {\alpha_i} = \frac{\exp(a_i)}{\sum_k \exp(a_k)}
    \end{aligned}   \label{eq:2}
\end{equation}
\noindent After obtaining the normalized matching scores, referred to as attention weights, we create a weighted sum of the context vectors. This weighted sum (Equation \ref{eq:3}) represents the attended information from the set of context vectors ${y_i}$ with respect to a specific query vector $x$.
\begin{equation}
    Att_{X \to Y}(x,\{y_i\}) = \sum_i \alpha_i y_i  \label{eq:3}
\end{equation}
Self-attention is when the \emph{query} vector $x$ is in the $\{y_j\}$ set of \emph{context} vectors. Although we can have the \emph{query} and \emph{context} vectors from mutually exclusive sets and can retrieve information from different domains (e.g., vision and language) by bringing them to a common vector space, which is how cross-modal attention is applied.

\subsubsection{Single Input Self Attention Encoders} The pair of single-input multi-headed self-attention (MSA) encoders in the framework operates on the offline extracted appearance, motion, and linguistic features (questions, answer candidates). At a particular hierarchical level of the complete pass of the framework, one input is an independent modality input, while the other can be another modality input or the output of a cross-fusion from a previous hierarchical level as in Figure \ref{fig:1}. 
Unlike the standard implementations with SA applied only to language \cite{devlin2019bert}, vision \cite{tan2019lxmert}, or any single modality input, our inputs are single modality as well as some previously fused semantics-aligned embeddings (Described in Cross Fusion Section). 
Each single-input self-attention encoder is built of a self-attention sublayer followed by a feed-forward sublayer with a residual connection and layer normalization added after each sublayer, following \cite{NIPS2017_3f5ee243}. 

\subsubsection{Cross Encoder} The cross-modality encoder consists of pair of linear layers $f_j$ and $g_j$ on both ends of the multi-headed cross-attention ($MCA$) sublayer, where $f_i$ and $g_i$ are acting as projection and back-projection functions mainly for dimension alignment and we apply a \emph{Pre-LayerNorm} ($LN$) on the inputs ($X_{M_1}^{SA}$ and $X_{M_2}^{SA}$) to multi-head cross attention ($MCA$) function and is finally followed by a residual connection.
\begin{equation}
    \begin{aligned}
    X_{M_1}^{CA}, X_{M_2}^{CA} = MCA\bigl([f_{M_1}&(LN(X_{M_1}^{SA})), \\ 
    &f_{M_2}(LN(X_{M_2}^{SA}))]\bigr)  \\
    X_{M_j}^{k+1} = X_{M_j}^k + g_j(&X_{M_j}^{CA})
    \end{aligned}   \label{eq:4}
\end{equation}
As mentioned in the background, the query and context can be mutually exclusive, and information can be retrieved, keeping any set as context using the other as a query and this flexibility is leveraged in the cross-encoders. This is shown by Equation \ref{eq:5},\ref{eq:6},\ref{eq:7}:

\begin{equation}
    \begin{gathered}
        K_M = X_{M}^{SA} \cdot W_{k} \quad Q_M = X_{M}^{SA} \cdot W_{q} \\
        V_{M_j} = X_{M_j}^{SA} \cdot W_{v_j} \label{eq:5}
    \end{gathered}
\end{equation}
\begin{equation}
    \begin{aligned}
        A_{M_1 \to M_2} = softmax(\frac{Q_{M_1}K_{M_2}^T}{\sqrt{d/h}})  \\
        A_{M_2 \to M_1} = softmax(\frac{K_{M_2}Q_{M_1}^T}{\sqrt{d/h}})
    \end{aligned}   \label{eq:6}
\end{equation}
\begin{equation}
    X_{M_1}^{CA} = A_{M_1 \to M_2} \cdot V_{M_2}    \label{eq:7}
\end{equation}


$W_k$, $W_q$, $W_{v_j}$ are the parameter matrices and $d$ is the inner dimension of cross-attention layer.
The overall encoder system has $K$ layers (refer Figure \ref{fig:2}) and the output of the $k$-th cross-modality encoder will again be given as input to the SA encoder at the $(k+1)$-th layer.

\subsection{Distraction Removal}
We have termed the process of removing irrelevant information from vectors as \emph{“distraction removal”}. When humans fail to ignore task-irrelevant or task-relevant distractions, it can interfere with their ability to complete tasks effectively \cite{Forster2008FailuresTI}. For instance, such distractions can lead to dangerous situations like car accidents\cite{ARTHUR199273}. 
The hypothesis is that incorporating distractions into the data, even with a small weight as in attention \cite{bahdanau2016neural}, negatively affects the model's training process and prediction accuracy during inference. Intuitively, even a small amount of irrelevant information can deteriorate the model's performance, akin to how distractions can influence human performance.


We first formulate a simple method to identify distractions and then describe the removal mechanism in a multi-headed attention setting. This eventually enhances our model’s focus on task-relevant information among the cross-information interactions. 
\begin{figure}[h!]
  \centering
  \includegraphics[width=0.9\columnwidth]{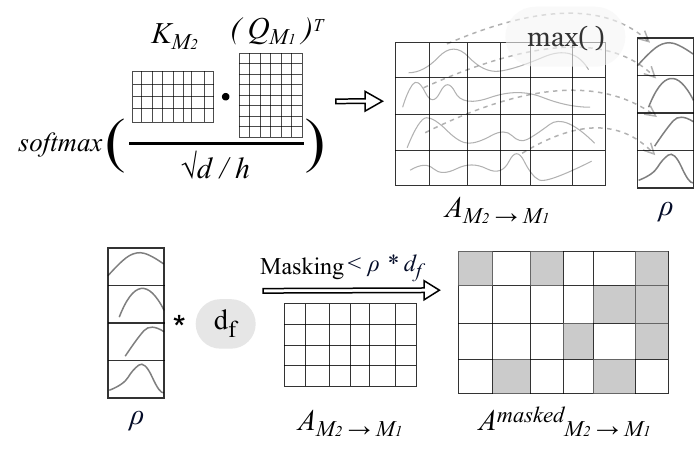}
  \caption{(Upper) Shows representative \emph{relevance score} $\rho$ generation from the CA matrix. (Lower) $\rho$ * $d_f$ sets the threshold to determine and mask out distracting weights in $A_{M_2 \to M_1}$}  \label{fig:3}
\end{figure}
\subsubsection{Distraction Identification} The result of taking the dot product of the projected \textit{embeddings matrices}, $Q$ and $K$, and then normalizing the scores using $softmax$ yields the attention weights matrix as shown in Equation (\ref{eq:6}). In this matrix, each row represents the ``relevance" of the context vectors $y_j$ (columns) for each of the query vectors $x$ (rows).
\begin{equation}
    \rho = max(A_{M_1 -> M_2}, dim=-1)
\end{equation}
The highest \emph{relevance} w.r.t. each query (highest in each row) becomes the representative \emph{relevance} score $\rho$ (for the query), and a threshold $\tau$ is set as a percentage of $\rho$ multiplying by \emph{distraction factor} $d_f$ (e.g., $0.3: 30\%$ of $\rho$). All attention weights below $\tau$ are considered \emph{distractors}.
\begin{equation}
    \tau = \rho * d_f
\end{equation}
\subsubsection{Distraction Removal in Multihead Subspaces}
We have adopted the multi-head attention mechanism as described by \cite{NIPS2017_3f5ee243}, having its essence in the distraction removal process as well. 
The core concept behind the multihead attention mechanism involves dividing the input embedding into $h$ different subspaces, also known as heads. Each head is responsible for learning specific relationships and patterns within its subspace. By calculating joint attention across these different subspaces, the model can capture cross-relations or information from various parts of the embedding.
This approach has the advantage of identifying distractor subspaces within the representations rather than considering the entire embedding vector as a distractor. By doing so, the model can identify irrelevant information more precisely in specific parts of the vectors, making the distraction removal process more effective. The multi-head attention mechanism helps localize and handle distractions more efficiently.

\begin{figure}[h!]
  \centering
  \includegraphics[width=.5\columnwidth]{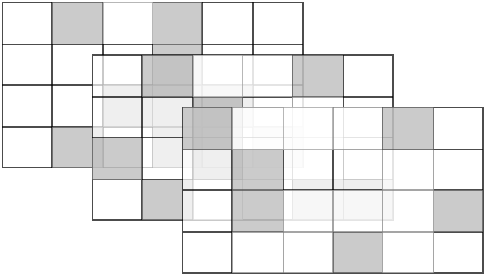}
  \caption{Different parts (subspaces) of the embeddings masked due to multiple heads. Above example: $3$-head space ($H=3$)}
\end{figure}



The distraction removal is then applied to these joint attention weights, and the context vector heads with attention weights below $\tau$ are set to $0$ in the attention weight matrix $A_{M_1 -> M_2}$ before the weighted averaging process (Equation \ref{eq:3}) to generate attended representations, which, in implementation is a modified Equation (\ref{eq:7}): $A_{M_1 -> M_2}^{masked}\cdot V_{M2}$ (Figure \ref{fig:3}).
\begin{equation}
  \mu_i = \Bigl\{ \bigl\{ a_{i,j}^h < \tau_i^h \bigr\}_{j=1}^{len(X_{M_2})} | a_i^h \in A_{M_1 -> M_2}^h \Bigr\}_{h=1}^H
\end{equation}
\begin{equation}
    (a_{i,j}^h)^{masked} = a_{i,j}^h * (1 - \mu_{i,j}^h)
\end{equation}
\noindent Here $\mu$ is the boolean mask for the weights in $A_{M_1 -> M_2}$ to be set to $0$, $H$ is the total number of heads and $len(X_{M_2})$ gives the number of row vectors in $X_{M_2}$ which equals number of columns in $A_{M_1 -> M_2}$ (i.e., context vectors \emph{relevance scores}).

\smallskip
\subsubsection{Distraction Factor}
In addition to learning more complex representations and learning to attend to different combinations of attention weights of the input and output sequences, the repetition of the overall encoder system in our framework also refines the distraction removal process. 
At each of the $K$ layers of the encoder system, the \emph{distraction factor} $d_f$ is increased by some value determined by a hyperparameter $\delta$. 
\begin{equation}
    d_f^{k+1} = d_f^{k} + \delta
\end{equation}
This makes the threshold more strict to distractors, and by the last layer, only the most relevant \emph{context} vectors  with high attention weights would be taken for the weighted averaging to generate an attended vector for a \emph{query} $x$. Intuitively, this can be seen as enhancing the relevant semantic information in the embeddings.
\begin{figure}[h!]
  \centering
  \includegraphics[width=\columnwidth]{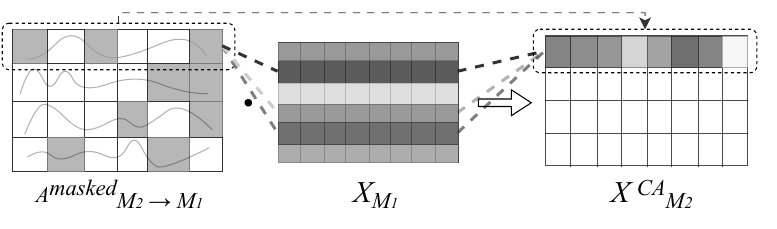}
  \caption{Masked (grey boxes in left figure) or $0$ weighted positions are ignored during the dot product in multi-head space. (Right most) Figure shows a weighted average of only $2^{nd}$, $4^{th}$, and $5^{th}$ embedding vectors. This also shows vector space transformation (in cross fusion) as the semantic alignment process is shown for $X_{M_1}$ vector being transformed to the space of $X_{M_2}$.}
\end{figure}
\subsection{Cross-Aligned Fusion}   \label{sec:crossfusion}
Previously, the generation of fused multi-modal representations has been based on simple concatenation and linear projection of vectors \cite{Xu_2021_CVPR} or parallel concatenation of the tokens with [CLS] token and applying cross-attention \cite{tan2019lxmert, lei2021less}, following \cite{devlin2019bert}, simply using the CLS as the cross-modality representation. 

We apply a semantic and dimensional alignment of vectors as a part of the DRAX instead of simply concatenating and linear projection or using the CLS token. We also implement the distraction removal operation in the vector space projection stage for semantically aligning only the relevant and non-distraction features. 

\subsubsection{Vector Space Transformation} We’re using the original interpretation of attention for aligning the \emph{important} vectors \cite{bahdanau2016neural} from a different (\emph{tailing}) vector space $X_t \in \mathbb{R}^{m \times d}$ to build the \emph{context} for particular \emph{query} tokens of the \emph{anchor} vector space $X_a \in \mathbb{R}^{n \times d}$. 
The $X_t$ vectors undergo a vector space transformation for the semantic as well as dimension alignment with the $X_a$ vector space.
\begin{equation}
    A_{a \to t}^{masked} = DR\bigl(softmax(\frac{Q_{a}K_{t}^T}{\sqrt{d/h}}) \bigr)  \label{eq:13} \\
\end{equation}
\begin{equation}
    X_t^{align} = A_{a \to t}^{masked} \cdot X_t    \label{eq:14}
\end{equation}
Attention matrix $A_{a \to t}^{masked}$ acts as the $\mathbb{R}^{m \times d} \to \mathbb{R}^{n \times d}$ space transformation or alignment matrix for $X_t$. Each row in $A_{a \to t}^{masked}$ would contain constants to form the linear combinations for the new rows of $X_t^{align}$. Although $A_{a \to t}^{masked}$ also undergoes \emph{distraction removal} $DR$ (as in the Distraction Removal section), which sets some $\alpha_i(s)$ to $0$ hence leaving out some parts (\emph{subspaces}, due to multi-head attention) of vectors. In the new $X_t^{align} \in \mathbb{R}^{n\times d}$ space, each vector is just a linear combination (${\alpha_1}x_1 + {\alpha_2}x_2 + ...$) of the original $X_t$ vectors (Equation \ref{eq:3}), but due to \emph{distraction removal}, some irrelevant parts of vectors are lost which introduces changes in the vector space \cite{strang1976linear}.
\subsubsection{Anchor Vectors} While fusing two sets of vectors, a decision is involved regarding which vector matrix is to be chosen as the anchor $X_a$. Meaning that the dimensions of one of the vector-matrix $X_t$ would be changed according to the other (anchor) to \emph{align} dimensions for the concatenation and fusion operation. The best resulting (empirical) combination is reported in the main results of the experiments section, while the ablation is shown further.

The semantically aligned $X_t^{align}$ vector is finally concatenated (along $dim=-1$) with $X_a$ and undergoes a feed forward fusion operation:
\begin{equation}
    X_{fused} = [X_a || X_t^{align}] \cdot W_f + b
\end{equation}
\subsection{Answer Decoder}
As shown in Figure \ref{fig:1}, after the question conditioning, the embeddings are repeated for the number of answer candidates $|\mathcal{A}|$, (along the batch dimension)\footnote{Implementation insight} so we can fuse the relevant information into the space of individual answer candidates, again, using the \emph{DRAX} block. Until now, CLS token-removed embeddings were passed between hierarchical blocks after fusion. But here, to get a final representation for every answer candidate $X_{fused}^{answer} \in \mathbb{R}^{4 \times d}$, we take a mean along with the CLS, of the final vectors reducing $\mathbb{R}^{4 \times(m\times d)}\to\mathbb{R}^{4 \times d}$, which is passed through the classifier to get the final label probabilities $p \in R^{|\mathcal{A}|}$. 
\begin{gather}
    y = ELU(W_a X_{fused}^{answer} + b) \\
    y' = ELU (W_y y + b) \\
    p = softmax (Wy'y' + b)
\end{gather}

\noindent Following \cite{le2020hierarchical} we use the hinge loss \cite{jang2017tgif} on pairwise comparisons, $max(0,1+ n - p)$, for incorrect $n$ and correct $p$ answers.



\section{Experiments}
We compare with related frameworks previously known with offline extracted feature vectors. We first discuss the main results and then the ablation of components, along with a discussion about the capabilities of the model tested by each task. Reported results are averaged over $3$ different seeds along with the standard deviation ($\sigma$). See Appendix B for visualizations of samples from the dataset.

\subsection{Main Results}
Table \ref{tab:main_results}  displays our framework outperforming previous similar SOTA approaches that operated on offline extracted features. We get an accuracy of $40.4$ ($\sigma = 0.76$) with a gain of $3.35$ ($9.04\%$) and $3.91$ ($10.71\%$) absolute (and relative) scores over \emph{ECLIPSE} \cite{Xu_2021_CVPR} and \emph{HCRN} \cite{le2020hierarchical}, respectively, with our large model having $K=6$ encoder layers, i.e., DRAX-large. 
Although for ablation study, we report the most significant results with the following parameters: $3$ encoder blocks (i.e., DRAX-base), an initial masking factor $d_f$ of $0.3$ (i.e., 30\% of ``representative relevance" score) for cross-attention masking, which increases by the same factor $\delta = 0.3$ ($30\% \to 60\% \to 90\%$) at each encoder layer, and a fusion masking factor of $0.4$, which stays constant as there is only a single fusion block. Encoder layers are reduced to $3$ due to memory and compute time constraints. 


\begin{table}
\begin{adjustbox}{max width=1.0\textwidth}
\renewcommand{\arraystretch}{1.3}
\setlength\tabcolsep{1.2pt}
\noindent\begin{minipage}{\linewidth}
\centering
\begin{tblr}{
  cells = {c},
  hline{1-2,15-16} = {-}{},
}
Models & Accuracy\\
Q-type (random) & $25.00$\\
QE-LSTM & $25.21$\\
QA-LSTM & $26.65$\\
Avgpooling & $30.45$\\
CNN+LSTM & $30.78$\\
I3D+LSTM & $33.21$\\
VIS+LSTM \cite{NIPS2015_831c2f88} & $29.91$ \\
BERT-VQA \cite{Yang_2020_WACV} & $33.68$ \\
TVQA \cite{lei-etal-2018-tvqa} & $35.16$\\
HCRN \cite{le2020hierarchical} & $36.49$\\
Eclipse \cite{Xu_2021_CVPR} & $37.05$\\
\textbf{DRAX}-base (ours) & $39.63$\\
\textbf{DRAX}-large (ours) & $\textbf{40.4}$\\
\emph{Human} & $95.43$
\end{tblr}
\captionof{table}{DRAX-base and large comparison with previous methods}\label{tab:main_results}
\end{minipage}
\end{adjustbox}
\end{table}

\begin{table*}[ht!]
\noindent\begin{minipage}{\linewidth}
\begin{adjustbox}{max width=\textwidth}
\renewcommand{\arraystretch}{1.3}
\setlength\tabcolsep{1.2pt}
\begin{tblr}{
  cells = {c},
  cell{1}{3} = {c=6}{},
  hline{1-3,8} = {-}{},
  hline{5} = {-}{dashed},
}
\textbf{Method} &  & \textbf{Task Accuracies} &  &  &  &  & \\
 & \textbf{Full} & {Basic\\Understanding} & Attribution & {Event\\Forecasting} & {Reverse\\Reasoning} & {Counterfactual\\Inference} & Introspection\\
HCRN & $36.4 \, (0.6)$ & $38.0 \, (0.33)$ & $33.59 \, (1.6)$ & $28.99 \, (1.12)$ & $29.65 \, (0.38)$ & $0.4 \, (0.54)$ & $0.2477 \, (2.72)$\\
Ours & $\textbf{39.63 \, (0.24)}$ & $37.48 \, (0.5)$ & $38.58 \, (0.59)$ & $30.94 \, (0.45)$ & $35.34 \, (1.01)$ & $43.06 \, (1.89)$ & $31.76 \, (0.67)$\\
{Ours $-$\\Cross(X)-Aligned Fusion} & $38.75 \, (0.14)$ & $37.19 \, (0.27)$ & $36.9 \, (0.06)$ & $29.59 \, (3.5)$ & $35.0 \, (2.78)$ & $41.98 \, (1.55)$ & $29.73 \, (4.78)$\\
{Ours $-$\\Distraction(D) Masking} & $38.93 \, (0.33)$ & $37.4 \, (0.47)$ & $39.24 \, (0.49)$ & $31.98 \, (1.81)$ & $37.45 \, (2.65)$ & $41.98 \, (1.64)$ & $35.14 \, (0.1)$\\
{Ours $-$ \\{[X-Alignment \& D-Masking]}} & $38.06 \, (0.2)$ & $38.04 \, (1.05)$ & $38.58 \, (0.26)$ & $29.14 \, (1.26)$ & $35.45 \, (0.1)$ & $42.97 \, (1.14)$ & $33.78 \, (0.1)$
\end{tblr}
\end{adjustbox}
\captionof{table}{Comparison with baseline HCRN and ablation of components for full dataset and sub-tasks upon DRAX-base. Accuracies mentioned are averaged on 3 seeds, ``$( \; )$" marks standard deviation and ``$-$" represents the removal of component(s).}\label{tab:2}
\end{minipage}
\end{table*}

\subsection{Tasks}

Table \ref{tab:2} shows task-wise accuracies. We discuss the diverse abilities of the framework tested by the tasks below:  

\subsubsection{Basic Understanding} This task tests a basic-level perception of the model. It covers queries regarding existing features in the scenes (e.g., types of objects, vehicles, and environment situations) and events or event classification (e.g., if an accident happened, type of accident, and actions of the pedestrians). Our method shows comparable performance to the HCRN. This task consists of the largest subset of the whole dataset with very basic queries, and the results show that distraction removal or semantic alignment doesn't play a significant role in solving this subset individually.

\subsubsection{Attribution} We focus on the model's capabilities for causal analysis in this task (e.x., what are the reasons for this crash). We get a score of $38.58$ with an increase of $4.89$ absolute points over the baseline and more robustness in results with a standard deviation ($\sigma$) of $0.59$ versus $1.6$ in the baseline.

\subsubsection{Event Forecasting} Testing the model's ability to predict future events by observing and analyzing a given scene, we are producing a score of $30.94$ with $\sigma = 0.45$, gaining $1.95$ over the baseline ($\sigma = 1.12$).

\subsubsection{Reverse Reasoning} This task makes the model look into the past of the provided segment of scenes and answer queries. Our method gets a score of 35.34 ($\sigma = 1.01$) with a gain of $5.69$ points over baseline.

\subsubsection{Counterfactual Inference} This tests the model's understanding of hypothetical situations in the context of the videos (e.g., would the accident still occur if the driver slows down in time?). So the model has to make inferences based on the imagined situations under given conditions. Our method scores $43.06$ gaining $3.06$ points over the baseline. 

\subsubsection{Introspection} This lets the model learn to provide preventive advice for avoiding certain traffic situations (e.g., Could the vehicle steer away to prevent the accident?), which actually tests the model's capabilities to think and provide resolutions. We get a score of $31.76$, gaining $6.99$ over baseline with a significant improvement in standard deviation $\sigma_{HCRN} = 2.72 \to \sigma_{DRAX} = 0.67$.

\subsection{Ablation Study on Full Dataset}

\subsubsection{Cross-Aligned Fusion} We ablated the component, replacing it with a simpler concatenation and linear projection fusion operation similar to HCRN. For appearance-motion feature fusion, we average consecutive 16 frames corresponding to each of the 8 clips, concatenate them along hidden dimension ($d$), and take a linear projection through a linear layer. Instead of using the final hidden state of $LSTM$ layer for the question tokens as in \emph{HCRN} and \emph{Eclipse}, we use the CLS token after self-encoder to repeat and concatenate with the previously fused embedding. This shows a decrease in the performance by $0.88$ on the full dataset and a subsequent performance decrease in subset tasks.

\subsubsection{Distraction Masking} Removing distraction masking brought down the result by $0.7$ points for the full dataset. Not getting a similar decrease in subtasks, we infer that distraction masking is beneficial for larger and more generalized dataset.

\subsubsection{Removing Distraction Masking and Cross-Alignment} A decrease of $1.57$ points is seen by completely ablating our proposed methods which clearly displays their significance in cross attended ``\emph{distraction-free}" embeddings.
\begin{table}[h!]
\begin{minipage}{\columnwidth}
\centering
\begin{tblr}{
  column{1} = {c},
  column{2} = {c},
  hline{1-2,5} = {1-2}{},
}
{Projections} & Accuracy ($\sigma$) & \\
{Appearance $\to$Motion$\to$Question\\ $\to$Answer}& $38.84$ ($0.61$) & \\
{Appearance$\to$Motion$\to$Question\\ $\gets$Answer}& $38.47$ ($0.75$) & \\
{Appearance$\to$Motion$\gets$Question\\ $\gets$Answer}& $38.74$ ($0.23$) & 
\end{tblr}
\captionof{table}{Anchor Space Projection Directions}\label{tab:vector_spaces}
\end{minipage}
\end{table}
\subsubsection{Anchor Vector Spaces} \quad Our fusion mechanism includes vector space projection from a tailing to anchor space. The best results were achieved with \{Appearance $\to$Motion$\gets$Question$\to$Answer\} where the ``$\to$" and ``$\gets$" notations denote direction of space projection. As our framework has a hierarchical structure, \{Appearance $\to$Motion$\gets$Question\} means appearance feature vectors are projected to motion vector space, and then question feature vectors are projected to the previous fused vectors' space. Experiments on the rest are shown in Table \ref{tab:vector_spaces}

\subsubsection{Using [CLS] for decoding} As in the more recent works like \emph{ClipBERT} \cite{lei2021less}, \emph{BLIP-2} \cite{li2023blip2} and \emph{VIOLET} \cite{fu2021violet}, we've experimented using only the final [CLS] embedding for decoding which gives a score of $38.47$ ($0.55$) showing that instead, taking a mean of our semantically aligned embeddings provides more significant query related information.

\section{Conclusion}

We presented our novel framework DRAX with the goal of producing ``distraction-free" and semantically aligned embeddings from cross-modality interactions. Instead of adding extra modalities, refining the input information (e.g., tokens), or heavy pre-training of the model before applying it to a task, we simply rid our embeddings of distractions in the latent space. 
As has been explicitly mentioned above, comparing larger models like BLIP-2, VIOLET, and ClipBERT is away from the scope of our study, and we've focused only on related previous works. 
DRAX demonstrates the existence of distractors in the embeddings and the advantage in removing them.
Applying the distraction removal mechanism on other video-language understanding tasks is the clear next step to our work, and we encourage the reader to do it before we do.

\bibliography{aaai24}
\clearpage
\appendix

\section{Appendix A}

\subsection{Implementation Details}

Each of our videos is divided into 8 clips, and 16 frames are sampled out of it. We apply the ResNet-18 model to extract the appearance features from the 128 frames per video, which are obtained from the penultimate layer of the model. They produce 512-dimensional vectors for each of the frames.
The 8 clips were used to compute the motion features, which include the temporal context of the clips into the features, producing 2048-dimensional vectors.

For linguistic inputs, i.e., questions and answer candidates, word embeddings are produced for each of their word tokens using the GloVe embeddings producing 300-dimensional vectors corresponding to each word.

Our models were implemented on a single NVIDIA A100 with 40GB GPU memory on an Ubuntu 20.04 operating system with Python 3.9.13 and PyTorch version 2.0.1. 

\subsubsection{Using the CLS}

As described in the paper, we add a CLS token with the embedding vectors before they are fed to the first stage of DRAX, i.e., the self-encoders. But, before the finally fused vectors are passed to the next state (Appearance-motion DRAX $\to$ Question DRAX $\to$ Answer Candidate DRAX), these CLS tokens are removed from the representations. In between the DRAX blocks, they help in sharing the overall information with each of the other token embeddings during the self and cross-attention operations. But after the final answer candidate fusion, instead of just using them as the output embeddings, they are just kept for a mean operation on the embeddings to get a single representation for each answer candidate to be fed to the classifier. 

\subsection{Dataset Description}

The SUTD TrafficQA dataset, consisting of 62,535 Question-Answer pairs and 10,080 videos, stands out as a prominent and widely recognized dataset not only for driving scene analysis but as a strong benchmark for VQA systems. This is because the dataset encompasses a diverse range of scenarios, presenting six challenging reasoning tasks for comprehensive evaluation and analysis:

\subsubsection{Basic Understanding} Contains a total of 38,773 queries, which require and test the model’s basic level perception and understanding of the traffic scene, including queries regarding features such as that of the vehicle and its type, the situation of the road, and the environment of the scene. 
Further, queries regarding the events and counting include events' temporal relation, analysis of influencing actions by the pedestrians, and classification of events into the type of accidents if an accident occurs at the scene.

Sample queries: 

\begin{itemize}
    \item Is there any traffic light violation?
    \item Is there any vehicle doing lane changing in the video?
\end{itemize}

\subsubsection{Reverse Reasoning} With a total of 2,769 queries, it requires a compound understanding of the scene for segment recognition while holding the relational information of moments temporally behind that segment. 

\begin{itemize}
    \item Did any pedestrians cross the road?
    \item What types of vehicles were involved in the accident?
\end{itemize}

\subsubsection{Counterfactual Inference} It has 3,009 queries creating hypothetical scenarios and requires the model to reason about the scenarios that did not actually occur in the video but is provided with conditions about the hypothetical situation. 

\begin{itemize}
    \item What types of vehicles that, if get removed from the videos, there won't be an accident?
    \item Would the accident still happen if all vehicles drive in their correct lane?
\end{itemize}

\subsubsection{Introspection} This type of query provides candidate answers for possible actions that could have been taken to avoid certain traffic scenarios like congestion or accidents and hence, train the model to provide preventive advice. The dataset contains 2,528 of these queries.

\begin{itemize}
    \item Could the vehicle steer away to prevent the accident?
    \item How can this road infrastructure be improved to prevent future accidents?
\end{itemize}

\subsubsection{Attribution} Attribution This is about learning the factors responsible for certain traffic scenarios, like analyzing reasons for an accident or the situations in which the accident happened, and has a total of 12,719 queries.

\begin{itemize}
    \item Where did the accident happen?
    \item Which factors might have contributed to the accident?
\end{itemize}

\subsubsection{Event Forecasting} Trains the model to make inferences regarding future events with a total of 2,737 queries.

\begin{itemize}
    \item Which vehicles are about to be involved in an accident?
    \item Judging by the speed of my vehicle, will my vehicle crash into other vehicles?
\end{itemize}

\newpage

\onecolumn 

\section{Appendix B}

\subsection{Functioning of appearance-motion conditioning}

\begin{figure}[h!]
  \centering
  \includegraphics[width=0.95\textwidth]{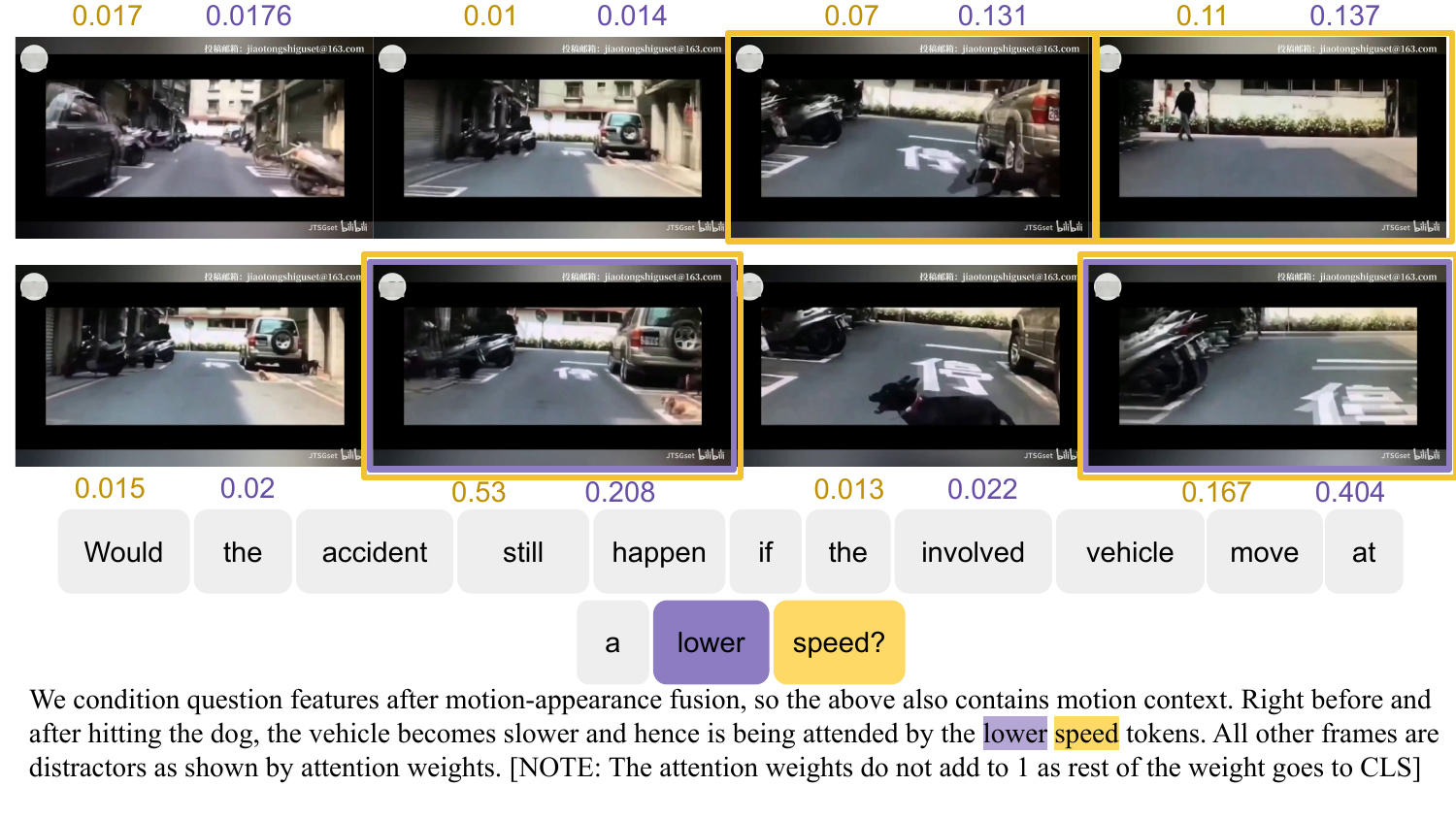}
  \caption{Video shows a vehicle colliding to a dog in first-person view. The video is repeated and shows the collision twice. Hence dog is visible in $3^{rd}$ and $7^{th}$ clips.}\label{fig:6} 
\end{figure}

Figure \ref{fig:6} is a clear example of how appearance and motion conditioning is functional. After the Appearance-Motion DRAX operation, the appearance frames are projected to the clip motion vector space and fused; the $8$ fused feature vectors would contain appearance as well as motion (temporal) context. This causes the ``lower speed" part of the question to attend to the collision part of the video as the vehicle slows down after hitting the dog.

\newpage

\subsection{Attentions and Distractions for question and answer tokens}

\begin{figure}[h!]
  \centering
  \includegraphics[width=0.95\textwidth]{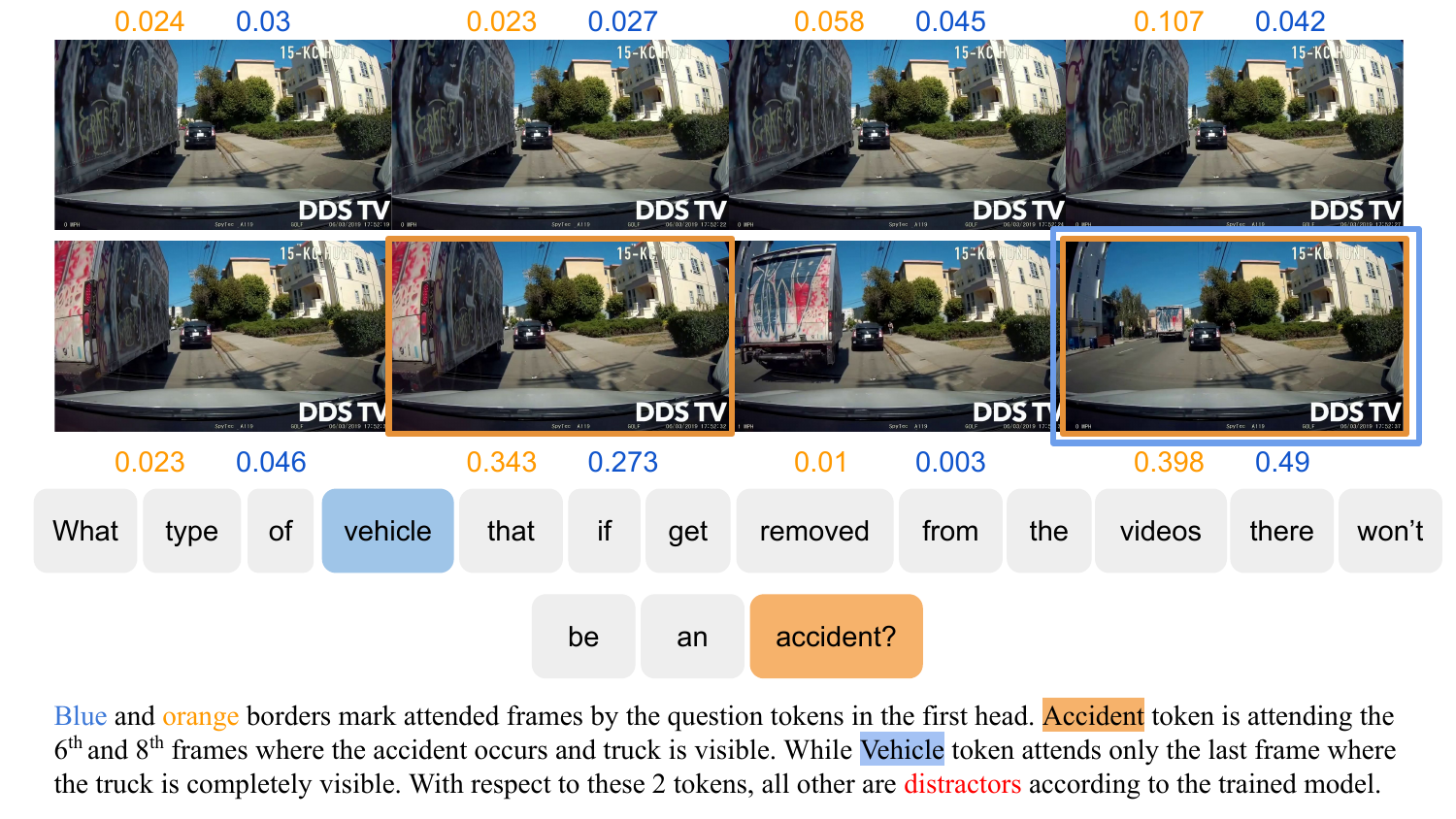}
  \caption{Scene of a heavy truck scraping through the bonet of }\label{fig:7} 
\end{figure}

Here (Figure \ref{fig:7} we show the token-clip attention between the question tokens, and each frame is annotated with the weights respective to tokens of focus. The frames not marked with the assigned colors are treated as distractors by the trained model during inference.

\begin{figure}[t!]
  \includegraphics[width=0.95\textwidth]{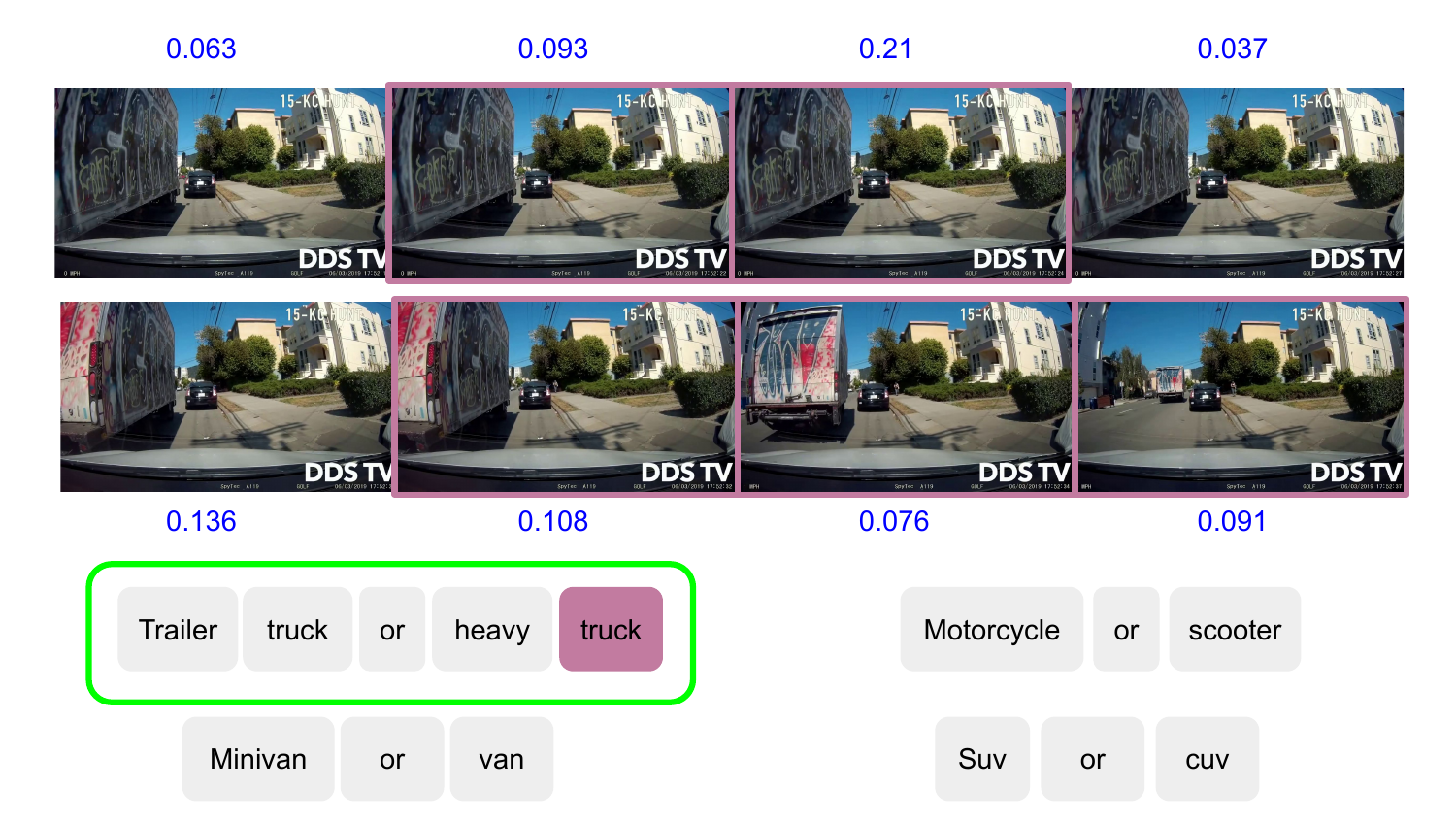} 
  \caption{Answer token attention for the video. Answer Candidate marked in green is the ground truth and predicted answer by the model.}\label{fig:8} 
\end{figure}

\end{document}